%% file: main_arxiv.tex
\documentclass{article}

\usepackage{styles/arxiv/arxiv}

\usepackage{graphicx}
\usepackage{amsmath}
\usepackage{float}
\usepackage[inline]{enumitem}
\usepackage{booktabs}
\usepackage{multirow}
\usepackage[]{algorithm2e}
\usepackage{tabularx}
\usepackage{caption}
\usepackage{subcaption}
\usepackage[pagebackref,breaklinks,colorlinks]{hyperref}
\usepackage{csvsimple}
\usepackage{amssymb}
\usepackage{bbold}

\begin{document}

    \title{Resource-Aware Neural Network Pruning Using Graph-based Reinforcement Learning}

    \input{text/authors/authors_arxiv}

    \maketitle

    \begin{abstract}
        \input{text/abstract}
    \end{abstract}

    \input{body}

    {
    \small
    \bibliographystyle{IEEEtran}
    \bibliography{references_filtered}
    }

\end{document}

%% file: text/authors/authors_arxiv.tex
\author{
    Dieter Balemans$^{1,2}$\thanks{dieter.balemans@uantwerpen.be}, 
    Thomas Huybrechts$^{1}$, 
    Jan Steckel$^{2,3}$, 
    Siegfried Mercelis$^{1}$ \\
    $^{1}$IDLab - Faculty of Applied Engineering, University of Antwerp - imec, Antwerp, Belgium \\
    $^{2}$CoSysLab - Faculty of Applied Engineering, University of Antwerp, Antwerp, Belgium \\
    $^{3}$Flanders Make Strategic Research Centre, Lommel, Belgium
}

%% file: text/abstract.tex
This paper presents a novel approach to neural network pruning by integrating a graph-based observation space into an AutoML framework to address the limitations of existing methods.
Traditional pruning approaches often depend on hand-crafted heuristics and local optimization perspectives, which can lead to suboptimal performance and inefficient pruning strategies.
Our framework transforms the pruning process by introducing a graph representation of the target neural network that captures complete topological relationships between layers and channels, replacing the limited layer-wise observation space with a global view of network structure.
The core innovations include a Graph Attention Network (GAT) encoder that processes the network's graph representation and generates a rich embedding.
Additionally, for the action space we transition from continuous pruning ratios to fine-grained binary action spaces which enables the agent to learn optimal channel importance criteria directly from data, moving away from predefined scoring functions.
These contributions are modelled within a Constrained Markov Decision Process (CMDP) framework, allowing the agent to make informed pruning decisions while adhering to resource constraints such as target compression rates. For this, we design a self-competition reward system that encourages the agent to outperform its previous best performance while satisfying the defined constraints.
We demonstrate the effectiveness of our approach through extensive experiments on benchmark datasets including CIFAR-10, CIFAR-100, and ImageNet.
The experiments show that our method consistently outperforms traditional pruning techniques, showing state-of-the-art results while learning task-specific pruning strategies that identify functionally redundant connections beyond simple weight magnitude considerations.

%% file: body.tex
\section{Introduction}\label{sec:intro}

\input{text/introduction}

\section{Methodology}\label{sec:methodology}
\input{text/methodology}

\section{Results}\label{sec:results}
\input{text/results}

\section{Conclusion and Future Work}\label{sec:conclusion-and-future-work}
\input{text/conclusion-and-future-work}

\section{Acknowledgments}\label{sec:acknowledgments}
This research was supported by the Research Foundation Flanders (FWO) under Grant Number 1SA8122N and 1SA8124N.

%% file: text/introduction.tex
Model compression has emerged as a critical technique for deploying deep neural networks (DNNs) in resource-constrained environments, such as mobile devices and edge IoT systems.
The increasing memory and computational demands of modern DNNs pose significant challenges for deployment in such settings, where hardware limitations and energy efficiency are paramount concerns.
Consequently, the field of neural network compression has gained substantial attention, with researchers focusing on reducing the computational complexity of DNNs while preserving their predictive performance.

Among the various compression techniques, pruning has established itself as one of the most effective approaches for model compression.
Pruning reduces network complexity by systematically removing redundant or less important parameters, thereby creating more efficient models suitable for deployment.
However, traditional pruning methods predominantly rely on hand-crafted policies that employ manually defined criteria to determine which network components should be removed.
As such, we identify two primary limitations of existing hand-crafted pruning methods.
First, these methods often adopt a local optimization perspective, focusing on individual layers or components of the network in isolation.
This layer-wise approach neglects the global structure and interdependencies within the network, potentially leading to suboptimal pruning decisions that fail to consider the entire network topology.
Second, these methods often depend on predefined scoring functions, such as weight magnitudes or activation statistics, to assess the importance of network parameters.
Magnitude-based pruning, for instance, assumes that weights with smaller absolute values are less important, but this assumption may not hold across different network architectures or application domains.
Furthermore, traditional methods typically rely on manually designed pruning policies determining both the locations and quantities of parameter removal.
This is inherently suboptimal and requires significant human expertise and time investment.
It can prove difficult to design effective pruning policies that adapt to the unique characteristics of each layer's contribution to the overall network performance.
Additionally, these policies depend mostly on predefined heuristic rules that may not be optimal for specific model-dataset combinations, potentially limiting their effectiveness across diverse applications.

To address these challenges, there has been growing interest in automating the pruning process through machine learning techniques.
Recent advances in AutoML \cite{Baratchi2024-es} for model compression have demonstrated promising results in overcoming the limitations of traditional hand-crafted approaches.
AutoML methodologies aim to automatically search for the best model architecture and hyperparameters without the need for human intervention.
Research by various teams \cite{Yang2018-dz, Li2020-yq, Liu2019-aa, Luo2018-qy, Yu2019-ar, Liu2020-gn} demonstrates that AutoML approaches often outperform traditional hand-crafted solutions.
Despite their heightened computational demands during training and optimization, the resulting models are typically more efficient and effective.
In the context of neural network pruning, AutoML methods often rely on reinforcement learning (RL) techniques to optimize the pruning policy of a structured pruning methodology \cite{He2018-uv, Feng2022-vg, Malik2021-oe, Lin2017-kw, Huang2018-cx}. 
Most notably, the work of He et al. \cite{He2018-uv} proposed the AutoML for Model Compression (AMC) framework, which employs a Deep Deterministic Policy Gradient (DDPG) reinforcement learning agent to learn optimal pruning schedules in a layer-wise manner.
This approach represents a significant advancement by automating the critical decision of how much to prune each layer, effectively replacing the labor-intensive manual trial-and-error process with a data-driven learned policy.

Despite these advancements, the AMC framework still exhibits fundamental limitations that constrain its effectiveness.
First, AMC continues to rely on predefined heuristic scoring functions, specifically the $l_1$-norm of weights, to determine the importance of individual channels or parameters.
This dependence on hand-crafted importance metrics undermines the automation goals of the framework and may not capture the true significance of network components for specific tasks.
Second, AMC operates with a limited observation space, utilizing only eleven hand-crafted features that focus exclusively on the current layer and basic compression statistics.
This layer-wise, local approach to pruning fails to capture the global network topology and the intricate interdependencies between layers and channels.
Consequently, while AMC represents a meaningful step toward automating the pruning schedule, it remains constrained by local optimization and predefined heuristics, potentially overlooking more effective compression strategies that would emerge from considering the network holistically.

To overcome the limitations of existing approaches, we propose a novel graph-based neural network pruning framework that fundamentally transforms the optimization process from local to global scope.
Our solution addresses both the dependence on heuristic scoring functions and the limited observation space by introducing a comprehensive graph representation of the neural network that captures the complete topological structure and inter-layer dependencies.

This is accomplished by transforming the target neural networks into graph representations.
This graph-based representation enables the preservation of both local layer-wise information and global network topology, providing a unified view of the network's structural relationships.
Unlike the layer-wise approach of AMC, this representation allows the optimization process to consider the impact of pruning decisions on the entire network simultaneously, leading to more informed and effective compression strategies.

To fully leverage this global perspective, we transition from hand-crafted heuristic scoring functions to a completely data-driven approach.
Our framework eliminates the reliance on predefined importance metrics such as $l_1$-norm weights, instead allowing the reinforcement learning agent to learn optimal importance criteria directly from the data.
This is achieved by reformulating the action space from continuous pruning ratios to fine-grained binary decisions, where the agent determines which individual channels to prune rather than specifying abstract compression rates for entire layers.
This binary action space provides the agent with precise control over the pruning process while maintaining the structured nature of the channel pruning compression.

The shift to binary actions and global observations necessitates a corresponding change in the reinforcement learning methodology.
We replace the Deep Deterministic Policy Gradient (DDPG) agent \cite{Silver2014-xw} used in AMC with a Proximal Policy Optimization (PPO) agent \cite{Schulman2017-ch}, which is better suited for discrete action spaces and provides more stable training dynamics.
Furthermore, we incorporate a novel self-competition reward system, inspired by the work of Mandhane et al. \cite{Mandhane2022-pn}, which has been successfully applied to video compression problems.
This reward mechanism incentivizes the agent to continuously improve upon its previous performance while explicitly prioritizing the satisfaction of resource constraints, addressing the implicit constraint handling limitations of the original AMC framework.

Similar to the work of Yu et al. \cite{Yu2021-so}, who employ a multi-stage graph embedding approach for network pruning, we leverage the graph-based representation to enhance the agent's understanding of the network structure.
Central to processing this representation is our Graph Attention Network (GAT) encoder \cite{Velickovic2017-ey}\cite{Brody2021-az}, which generates rich embeddings that capture both local node features and global structural patterns.
The GAT architecture is particularly well-suited for this task as it can dynamically attend to relevant network components, effectively learning which parts of the network topology are most important for making pruning decisions.
These embeddings serve as the observation space for the PPO agent, providing a comprehensive and learnable representation of the network's current compression state.

The entire framework operates in an end-to-end manner, with both the GAT encoder and PPO agent trained simultaneously through reinforcement learning.
This joint optimization allows the system to learn not only optimal pruning policies but also the most effective ways to represent and process network topology for compression purposes.

In summary, this paper presents several key contributions that advance the field of automated neural network compression:

\begin{itemize}
    \item \textbf{Graph-based global observation space}: We introduce a novel graph representation of neural networks that captures complete topological relationships between layers and channels, replacing the limited layer-wise observation space of existing methods with a comprehensive global view.
    
    \item \textbf{Data-driven importance learning}: Our framework eliminates dependence on hand-crafted heuristic scoring functions by enabling the agent to learn optimal channel importance criteria directly from data through fine-grained binary pruning decisions.
    
    \item \textbf{Graph Attention Network encoder}: We develop a GAT-based encoder that processes the graph representation to generate rich embeddings, replacing manually designed features with learnable representations that capture both local and global network properties.
    
    \item \textbf{Self-competition reward system}: We adapt the self-competition mechanism from video compression to neural network pruning, providing explicit incentivization for meeting resource constraints while improving upon past performance.
    
    \item \textbf{End-to-end learning framework}: The complete system, comprising PPO agent and GAT encoder, is trained jointly to optimize both network representation and pruning policy simultaneously, enabling adaptation to diverse architectures without manual intervention.
\end{itemize}

These contributions collectively address the fundamental limitations of existing automated pruning methods, establishing a foundation for truly generalizable compression frameworks that can operate effectively across different neural network architectures and application domains.

%% file: text/methodology.tex
\subsection{Overview}

In this section, we present our approach to neural network compression using Graph Attention Network (GAT).
We propose a novel methodology that uses a channel-level graph representation of the target neural network with a GAT as feature extractor for a reinforcement learning (RL) agent.
Our methodology uses a constrained Markov decision process (CMDP) framework to formulate the pruning problem, which we describe in detail in the following subsections.

\subsection{Structured Pruning as CMDP}

\begin{figure*}[ht]
    \centering
    \includegraphics[width=0.95\textwidth]{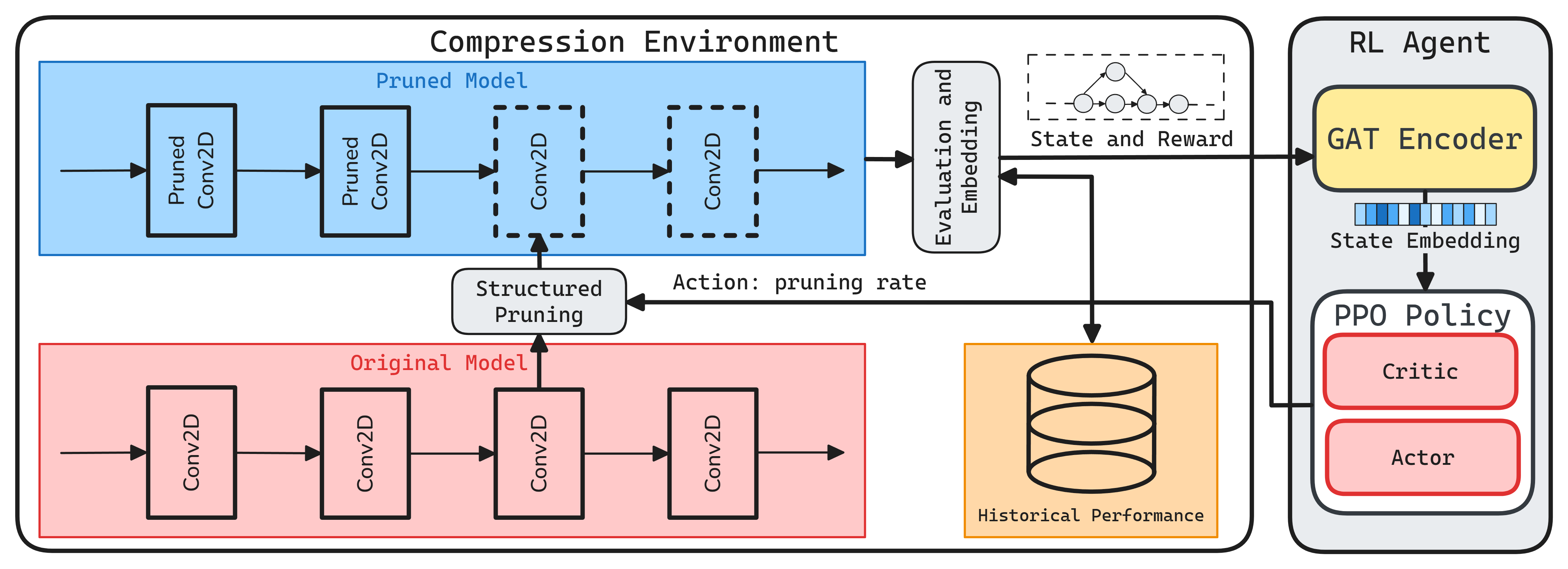} 
    \caption{Overview of the self-competition environment where the PPO agent interacts with the pruning environment. Each step, the agent takes an action that prunes part of the network. When the agent has finished pruning the entire network, it receives a reward based on whether its performance is better than the historical performance.}
    \label{fig:rl_overview}
\end{figure*} 

In this work, we follow the structured pruning environment presented in the original AMC paper \cite{He2018-uv}, with some modifications to adapt it to our graph-based approach.
In this environment, an agent interacts by taking pruning actions sequentially on the target network.
An overview of the environment is shown in Figure \ref{fig:rl_overview}.
At each step, the agent takes a pruning action that removes channels from the target neural network.
In the original AMC environment, the agent prunes one layer at a time, and an episode ends when all layers have been pruned.
However, in our approach, we use a channel-grouping method to allow the agent to prune the network in a more fine-grained manner.
For this, the target network is divided into equally sized groups of channels.
The episode ends when all groups of channels have been pruned.
Only then is the final network evaluated and the agent receives a reward based on its performance.
We model the pruning problem as a Constrained Markov Decision Process (CMDP) \cite{Altman2021-xf} and define the state space, action space, and reward space.
The goal of the environment is to maximize the accuracy of the target neural network while working within a resource constraint. 
The resource constraint can be defined by multiple factors, such as the number of floating-point operations (FLOPs), the number of parameters, memory footprint, or the latency of the network.
In this section, we use the number of FLOPs as the resource constraint, though the environment can be easily adapted to other constraints.
The CMDP tuple is defined as follows:
\begin{equation}
    \mathcal{M} = \langle \mathcal{O}, \mathcal{A}, \mathcal{P}, \mathcal{R}, \mathcal{C} \rangle
\end{equation}
where $\mathcal{O}$ is the observation space, $\mathcal{A}$ is the action space, $\mathcal{P}$ is the transition probability function, $\mathcal{R}$ is the reward function, and $\mathcal{C}$ is the constraint function.

\subsubsection{Observation Space}

The observation space $\mathcal{O}$ is defined as the graph representation of the target neural network.
The graph representation contains computed features of the elements of the target neural network that can inform the agent about the current state of the network.
The graph is structured as a directed acyclic graph (DAG), where each node represents a layer of the target neural network and each edge represents the connections between the layers.
Thus, the graph representation of the target neural network is defined as:
\begin{equation}
    \mathcal{O} = \langle \mathcal{G} \rangle = (\mathcal{V}, \mathcal{E})
\end{equation}
where $\mathcal{G}$ is the graph representation of the target neural network, $\mathcal{V}$ is the set of nodes representing the layers of the network, and $\mathcal{E}$ is the set of edges representing the connections between the layers.

For each node $v \in \mathcal{V}$, we define a feature vector $f_v \in \mathbb{R}^d$ that contains the computed features of the layer.
These features can be divided into three categories:
\begin{itemize}
    \item \textbf{Structural Features:} These features describe the structure of the layer, such as the number of input and output channels, the kernel size, and the stride.
    \item \textbf{Computational Features:} These features describe the computational cost of the layer, such as the number of floating-point operations (FLOPs), the number of parameters, and the memory footprint.
    \item \textbf{Weight Features:} These features describe the learned weights of the layer. We include the $l_1$-norm of each of the individual channels in the layer, which can be used to determine the importance of each channel. It is important to note that in order to have a valid feature vector, we need to ensure feature vector representation for each layer has a consistent dimension. This is achieved by padding the feature vector with zeros if the number of channels in the layer is less than the maximum number of channels in the target neural network.
\end{itemize}

For each edge $e \in \mathcal{E}$, we define a feature vector $f_e \in \mathbb{R}^d$ that contains the computed features of the connection between the layers.
The edge features are defined by the edge type, which can be a regular connection, a skip connection, or a residual connection.
Additionally, we include the $l_1$-norm of the of the output channel activations of the source layer, which can be used to determine the importance of the connection.

This results in a graph representation of the target neural network, where each node contains a feature vector that describes the layer and each edge contains a feature vector that describes the connection between the layers.
This representation preserves the topological structure of the target neural network and allows the agent to learn a pruning policy based on the features of the layers and connections.

\subsubsection{Action Space}

As mentioned before, we use a channel-grouping approach for the action space $\mathcal{A}$.
This differs from the original AMC environment, which uses a single continuous action space to prune one layer at a time.
This reformulation is needed as we adapt the action space to a fine-grained binary action mask, where each value in the mask corresponds to a channel in the target neural network.
To allow the agent to prune the network in a sequential manner, we define the action space as a set of channel groups. 
These groups could be defined as the number of channels in each individual layer, however, this would require action padding and masking to ensure that the action space is always the same size.
Therefore, we define the action space as a set of channel groups, where each group consists of a fixed number of channels.
This way we prune a single channel group at each step, allowing for sequential actions in the environment.
It is important to note that the group count can be configured, making the episode length configurable via this parameter.

The action space is defined as:
\begin{equation}
    \mathcal{A} = \{a_1, a_2, \ldots, a_{\frac{C}{n}}\}
\end{equation}
where $a_i$ is a binary action corresponding to pruning a channel $i$ from the current group of channels, $C$ is the total number of channels in the target neural network, and $n$ is the number of channel groups.

The parameter $n$ fundamentally determines the operational characteristics of our reinforcement learning framework, leading to two distinct paradigms for neural network pruning. 
When $n = 1$, the action space consists of a single group containing all channels of the target network, resulting in an episode length of exactly one step. 
In this configuration, the reinforcement learning agent functions as a combinatorial optimization algorithm rather than a traditional sequential decision-making system. 
The agent must simultaneously determine the optimal pruning configuration for the entire network in a single action, making this approach conceptually similar to traditional optimization algorithms such as genetic algorithms or evolutionary strategies \cite{Simon2013-fo}.

Conversely, when $n > 1$, the action space is partitioned into multiple channel groups, enabling the reinforcement learning agent to prune the network through a sequential decision-making process. 
This configuration represents reinforcement learning in its more traditional sense, where the agent learns a policy consisting of a trajectory of actions that collectively maximize the estimated reward in the environment. 
The agent must develop a sophisticated understanding of the state-action mapping for each group of channels, learning how early pruning decisions influence the optimal choices for subsequent groups. 
This sequential approach allows the agent to adapt its pruning strategy based on the evolving network structure, potentially discovering more nuanced pruning patterns that account for the interdependencies between different network components.
This approach also forces the agent to learn an optimal encoding using the graph representation of the target network, as the agent must learn to prune the network in a sequential manner. 
This can potentially lead to more informed pruning decisions, as the agent can leverage the learned features of the layers and connections to make better decisions about which channels to prune.

\subsubsection{Transition Probability Function}

The transition function $\mathcal{P}$ describes the probability of transitioning to the next state $s_{t+1}$ given the current state $s_t$ and action $a_t$, formally written as $\mathcal{P}(s_{t+1} | s_t, a_t)$.
In our pruning environment, the transitions are deterministic, meaning $\mathcal{P}(s_{t+1} | s_t, a_t) = 1$ for the actual next state and $0$ for all other states.
The transition function depends on the number of channel groups $n$ defined in the action space.
When $n = 1$, the action space consists of a single group containing all channels, and the transition function results in a terminal state after one step, as the agent prunes the entire target network at once.
When $n > 1$, the action space consists of multiple groups, and the transition function sequentially advances through each group.
In this setting, the next state $s_{t+1}$ is computed by updating the graph representation based on the pruning action $a_t$, where the corresponding nodes and edges are modified to reflect the reduced network structure, and the environment advances to focus on the next group of channels to be pruned.

\subsubsection{Reward Function and Constraint Function}

The reward function $\mathcal{R}$ is defined as the task performance of the target neural network.
The task performance can be measured using various metrics depending on the task itself.
In this work, we use the top-1 classification accuracy of the network.
The reward is only given at the end of an episode, after all layers have been pruned and the final network is evaluated.
Thus, we work with a sparse reward signal, which we take into account when training the RL agent.

The constraint function $\mathcal{C}$ is defined as a function depicting a desired constraint for the network.
In the compression environment, the agent is tasked with maximizing the accuracy of the target neural network while not exceeding a given resource constraint.
This resource constraint can be modeled using many different metrics, such as the number of FLOPs, the number of parameters, the memory footprint, or the latency of the network.
For example, we could set the target to 50\% of the original FLOPs of the network. 
In equation \ref{eq:constrained_objective}, we define the resource constraint as the number of FLOPs.
The global constrained objective function is defined as follows:
\begin{equation} 
    \max_{\pi(a_t \mid s_t, FLOPs_{\text{target}})} J_{\pi}^{\text{Acc}}
    \quad \text{s.t.} \quad J_{\pi}^{\text{FLOPs}} \leq FLOPs_{\text{target}}  \label{eq:constrained_objective}
\end{equation}
where $J_{\pi}^{\text{Acc}}$ is the expected sum of discounted rewards:
\begin{equation}
    \max_{\pi} J_{\pi}^{Acc}(s) := \mathbb{E}_{a \sim \pi, s \sim P} \left[ \sum_{t} \gamma^t Acc(s_t, a_t) \right]  \label{eq:accuracy_objective}
\end{equation} 
Here, $\gamma$ is the discount factor, set to either $0.0$ or $1.0$ depending on the action space groups ($n$).
The discount factor is set to $0.0$ when $n = 1$, as the agent prunes the entire network in a single step.
When $n > 1$, the discount factor is set to $1.0$, as the agent prunes the network in a sequential manner and the reward is only given at the end of the episode. By setting the discount factor to $1.0$, we ensure the expected sum of discounted rewards is the measured accuracy of the network after all layers have been pruned.

As mentioned in \cite{Mandhane2022-pn}, this compression environment can be approached as an unconstrained Lagrangian optimization problem.
However, as the authors argue, this approach is difficult to tune and most likely will not generalize to other environments.
Therefore, we follow their conclusion and propose using a self-competition mechanism.
This self-competition approach solves a slightly modified version of the described CMDP and removes the need for Lagrangian parameters.

The self-competition reward is designed for an agent to continuously improve upon its past performance.
The reward signal reflects the agent's performance with the goal of maximizing the accuracy while satisfying the given resource constraint, in this case the number of FLOPs.
The reward is given by:
\begin{align}
    \mathcal{R} &= 
    \begin{cases} 
      -sgn(F_{ep} - F_{EMA}) & \text{if } F_{ep} > F_{target} \\
      sgn(Acc_{ep} - Acc_{EMA}) & \text{otherwise}
    \end{cases}
\end{align}
where $Acc_{ep}$ is the accuracy of the current episode, $Acc_{EMA}$ is the exponential moving average of the accuracy, $F_{ep}$ is the number of FLOPs in the network in the current episode, $F_{EMA}$ is the exponential moving average of the number of FLOPs, and $F_{target}$ is the target number of FLOPs.
The sign function, denoted as $sgn$, outputs either $-1$ or $1$.
The agent receives a reward of $1$ if the current performance is better than the historical performance, and $-1$ if the current performance falls short.
The reward function operates like an if-statement, rewarding the agent based on either the current accuracy or the current FLOPs, depending on whether the target constraint is satisfied. 
Consequently, this reward function operates in two phases:
\begin{itemize}
    \item \textbf{Phase One:} The agent is rewarded for achieving a target number of FLOPs. The agent operates in this phase until the target number of FLOPs is reached. The reward is $-1$ if the pruned network's FLOPs are above the historical average, and $1$ if the pruned network's FLOPs are below the historical average.
    \item \textbf{Phase Two:} Once the target number of FLOPs is achieved, the agent is rewarded for maximizing the accuracy. The reward is $-1$ if the agent's accuracy is below the historical average, and $1$ if the agent's accuracy is above the historical average.
\end{itemize}

Finally, we argue that the self-competition reward system does not violate the Markovian property. 
As the agent strives to surpass its historical performance, this historical performance concurrently improves.
However, while it does learn from previous iterations, it does not inherently rely on a sequence of prior states within each individual scenario.
Therefore, the self-competition reward system aligns well with the Markovian property of the reinforcement learning framework.
Additionally, we can update the objective function to include the self-competition reward system:
\begin{equation}
    \begin{aligned}
    \max_{\pi(a_t \mid s_t, A_{\text{EMA}}, F_{\text{EMA}})} \quad & \text{sgn}(A_{\text{ep}} - A_{\text{EMA}}) \cdot \mathbb{1}_{F_{\text{EMA}} \leq 0, F_{\text{ep}} \leq 0} \\
    \text{s.t.} \quad & \text{sgn}(|F_{\text{ep}}| - |F_{\text{EMA}}|) \leq 0
    \end{aligned}
\end{equation}

where $A_{\text{ep}}$ is the accuracy of the current episode, $A_{\text{EMA}}$ is the exponential moving average of the accuracy, $F_{\text{ep}}$ is the number of FLOPs in the network in the current episode, and $F_{\text{EMA}}$ is the exponential moving average of the number of FLOPs.
The $\mathbb{1}$ function is an indicator function that outputs $1$ if the condition is true and $-1$ otherwise.

\subsubsection{Operation Modes of Self-Competition}

It is important to note that while we present the self-competition reward system with the number of FLOPs as the resource constraint, it can be easily adapted to other constraints.
For example, we can define different operation modes for the compression system, such as a resource constraint mode and a performance guaranteed mode.
In the resource constraint mode, the agent is tasked with maximizing the accuracy of the target neural network while not exceeding a given resource constraint.
This mode is presented in the previous sections.
The FLOPs metric could be replaced with other metrics, such as the number of parameters, the memory footprint, or the latency of the network depending on the needs of the user.
In the performance guaranteed mode, the agent is tasked with minimizing the resource consumption of the target neural network while not falling below a given performance threshold.
For the performance guaranteed mode, the reward function is defined as follows:
\begin{equation}
    \mathcal{R} = 
    \begin{cases} 
      sgn(Acc_{ep} - Acc_{EMA}) & \text{if } Acc_{ep} < Acc_{target} \\
      -sgn(F_{ep} - F_{EMA}) & \text{otherwise}
    \end{cases}
\end{equation}
where $Acc_{target}$ is the target accuracy of the network, $Acc_{ep}$ is the accuracy of the current episode, $F_{target}$ is the target number of FLOPs, and $F_{ep}$ is the number of FLOPs in the network in the current episode.
Using this reward function, the agent optimizes the models to minimize the number of FLOPs while ensuring that the accuracy of the network does not fall below a given threshold.
Again, the FLOPs metric could be replaced with other metrics.

\subsection{Graph-Based PPO Agent}

In this section, we present the graph-based PPO agent that interacts with the pruning environment.
The agent is based on the Proximal Policy Optimization (PPO) algorithm \cite{Schulman2017-ch}, which is a popular reinforcement learning algorithm that has been shown to be effective in a wide range of tasks.
PPO is an on-policy algorithm that performs conservative policy updates, resulting in more stable training.
In our case, for the structured pruning environment, we need an on-policy algorithm as the self-competition reward system provided non-stationary data and the agent tries to improve over its own past performance.
The agent will receive a reward based on its performance relative to its historical performance, which would not make sense in an off-policy setting. 
In an on-policy setting, the agent updates its policy based on the interactions it has with the environment based on its current policy only. 
This is in contrast to an off-policy setting, where the agent can learn from past experiences and update its policy based on those experiences.
In that setting, the agent would keep a large replay buffer of past experiences and sample from it to update its policy.
That replay buffer is not needed in our case. 

\subsubsection{GAT Encoder}

In order to process the graph representation of the target neural network, we use a Graph Attention Network (GAT) Encoder \cite{Velickovic2017-ey}\cite{Brody2021-az} as feature extractor for the agent.
This is shown in Figure \ref{fig:rl_overview}, where the GAT encoder processes the graph representation of the target neural network and outputs a feature vector for each node in the graph.

The encoder works by first generating updated node feature vectors for each node using the message passing operation of the GAT encoder.
The message passing operation of the GAT encoder is defined as follows:
\begin{equation}
    x'_i = \alpha_{i,i} \Theta_s x_i + \sum_{j \in N(i)} \alpha_{i,j} \Theta_t x_j  \label{eq:gat_operation}
\end{equation}
where $x'_i$ represents the updated feature vector for node $i$. The term $x_i$ denotes the input feature vector of node $i$, while $N(i)$ is the set of its direct neighbors. $\Theta_s$ and $\Theta_t$ are learnable weight matrices responsible for linearly transforming the features of the source node ($x_i$) and target nodes ($x_j$), respectively. The attention coefficient $\alpha_{i,j}$ quantifies the importance of node $j$'s features when updating node $i$'s representation.

These attention coefficients $\alpha_{i,j}$ are dynamically computed using an attention mechanism that incorporates the features of the interacting nodes and their connecting edges. The computation is detailed as:
\begin{equation}
    \alpha_{i,j} = \frac{\exp \left( \mathbf{a}^\mathsf{T} \mathrm{LeakyReLU} \left( \Theta_s \mathbf{x}_i + \Theta_t \mathbf{x}_j + \Theta_e \mathbf{e}_{i,j} \right) \right)}{\sum_{k \in N(i) \cup \{i\}} \exp \left( \mathbf{a}^\mathsf{T} \mathrm{LeakyReLU} \left( \Theta_s \mathbf{x}_i + \Theta_t \mathbf{x}_k + \Theta_e \mathbf{e}_{i,k} \right) \right)} \label{eq:gat_attention}
\end{equation} 
In this formula, $\mathbf{e}_{i,j}$ signifies the feature vector of the edge connecting nodes $i$ and $j$, transformed by the learnable weight matrix $\Theta_e$. The terms $\Theta_s \mathbf{x}_i$, $\Theta_t \mathbf{x}_j$, and $\Theta_e \mathbf{e}_{i,j}$ are summed to form a combined representation of the interaction between node $i$, node $j$, and their edge. This sum then passes through the LeakyReLU activation function, which introduces non-linearity. The resulting vector is then multiplied by a learnable attention vector $\mathbf{a}^\mathsf{T}$, which scores the interaction's relevance. Finally, the exponential function and subsequent normalization over all neighbors $k$ of node $i$ (including node $i$ itself) via a softmax operation ensures that the attention coefficients for each node sum to one.
This attention mechanism allows the GAT encoder to focus on the most relevant nodes and edges in the graph, enabling it to learn a more expressive representation of the target neural network.

The GAT encoder performs multiple message passing operations, where each operation updates the feature vectors of the nodes in the graph-based on their neighbors' features.
The number of message passing operations is configurable, allowing the agent to learn a more complex representation of the target neural network. 
We use a standard of three message passing operations.
After the message passing steps are completed, we have a new feature vector for each node in the graph.
This feature vector per node is defined as:
\begin{equation}
    e_i^{t} = \text{GATEncoder}(\mathcal{G}_t) \label{eq:gat_encoder_output}
\end{equation} 
where $e_i^{t}$ is the output embedding of a node $i$ at time step $t$ and $\mathcal{G}_t$ is the graph representation of the target neural network at time step $t$.

In order to provide a global embedding of the graph, we use a global attention aggregation mechanism \cite{Li2019-qx} to aggregate the node embeddings into a single global embedding.

This global embedding is defined as:
\begin{equation}
    e^{t} = \sum_{n=1}^{N_i} \mathrm{softmax} \left( h_{\mathrm{gate}}(\mathbf{x}_n) \right) \cdot h_{\Theta}(\mathbf{x}_n)  \label{eq:gat_pooling_output}
\end{equation}
where $e^{t}$ is the global embedding of the graph at time step $t$, $N_i$ is the number of nodes in the graph, $\mathbf{x}_n$ is the feature vector of node $n$, $h_{\mathrm{gate}}$ and $h_{\Theta}$ are learnable functions (MLPs) that compute the attention weights and the aggregated node features, respectively.

The global embedding is then used as the input to the policy and value networks of the PPO agent.

\subsubsection{Complete Agent Architecture}

The complete architecture of the PPO agent is shown in Figure \ref{fig:rl_overview}.
The agent consists of a GAT encoder that processes the graph representation of the target neural network and outputs a global embedding of the graph.
This global embedding is then used as the input to the policy and value networks of the PPO agent.
The policy network is a multi-layer perceptron (MLP) that takes the global embedding as input and outputs a probability distribution over the action space.
Since the action space is discrete and binary, we use the Bernoulli distribution to model the action probabilities.
The output of the policy network can be interpreted as logits, which are then transformed into probabilities using the softmax function.
The initialization of the policy network is of immense importance as it determines the initial action probabilities, and thus the initial pruning policy of the agent. 
Since we use a self-competition reward system, we want to agent to start with a lower sparsity and gradually increase it as it learns to prune the network.
Therefore, we initialize the last layers of the policy network with a bias term that push the action probabilities to be closer to 0, meaning that the agent will initially be more conservative in its pruning actions. 
Additionally, the policy network is initialized using Orthogonal Initialization \cite{Saxe2013-ua} with a small gain of 0.01, which helps to stabilize the training process and prevents the gradients from exploding or vanishing. 

Finally, PPO uses an actor-critic architecture, where the actor network is responsible for selecting actions based on the current state, and the critic network is responsible for estimating the value of the current state.

\subsubsection{Pre-training the GAT Encoder}

As mentioned before, the agent, including the GAT encoder, is trained end to end using the PPO algorithm.
However, training the GAT encoder from scratch can prove difficult, as the agent has to learn the graph representation of the target neural network and the pruning policy at the same time.
In order to improve the training stability, the GAT encoder can optionally be pre-trained in a self-supervised manner.
This pre-training is done using a Graph Autoencoder (GAE) \cite{Kipf2016-cx} that learns to reconstruct the graph representation of the target neural network.
This way, the GAT encoder has to learn an efficient encoding of the graph representation of the target neural network, which can be used to learn a pruning policy.
The idea is that the GAT encoder learns a highly separated informative embedding space, which the agent can then leverage to map the states to optimal actions.
The GAE architecture consists, first, of the encoder itself. 
Then, for each reconstruction objective we have a decoder head that reconstructs the graph representation of the target neural network.
The reconstruction objectives are:
\begin{itemize}
    \item \textbf{Adjacency Matrix Reconstruction:} The GAE reconstructs the adjacency matrix of the graph representation of the target neural network. This forces the GAT encoder to learn a meaningful representation of the topological structure of the target neural network.
    \item \textbf{Node Feature Reconstruction:} The GAE reconstructs the node features of the graph representation of the target neural network. This forces the GAT encoder to learn a meaningful representation of the features of the layers in the target neural network. This includes the structural, computational, and weight features of the layers.
    \item \textbf{Edge Feature Reconstruction:} The GAE reconstructs the edge features of the graph representation of the target neural network. This forces the GAT encoder to learn a meaningful representation of the connections between the layers in the target neural network. This includes the edge type and the $l_1$-norm of the output channel activations of the source layer.
\end{itemize}

The GAE network is then trained using a reconstruction loss. 
The reconstruction loss is defined as the sum of the reconstruction losses for each of the reconstruction objectives:
\begin{equation}
    \mathcal{L}_{\text{recon}} = \mathcal{L}_{\text{adj}} + \mathcal{L}_{\text{feat}} + \mathcal{L}_{\text{edge}}
\end{equation}
where $\mathcal{L}_{\text{adj}}$, $\mathcal{L}_{\text{feat}}$, and $\mathcal{L}_{\text{edge}}$ are the reconstruction losses for the adjacency matrix, node features, and edge features, respectively.
The reconstruction loss of the adjacency matrix is defined as the binary cross entropy loss \cite{Bridle1990-op}, while the reconstruction loss of the node features and edge features is defined as the mean squared error loss \cite{Rumelhart1986-db}.
The reconstruction losses are defined as follows:

\begin{align}
    \mathcal{L}_{\text{adj}} &= -\sum_{i,j} \left( A_{ij} \log(\hat{A}_{ij}) + (1 - A_{ij}) \log(1 - \hat{A}_{ij}) \right) \\
    \mathcal{L}_{\text{feat}} &= \sum_{i} (X_i - \hat{X}_i)^2 \\
    \mathcal{L}_{\text{edge}} &= \sum_{i,j} (E_{ij} - \hat{E}_{ij})^2   
\end{align}
where $A_{ij}$ is the entry in the adjacency matrix, $\hat{A}_{ij}$ is the predicted entry in the adjacency matrix, $X_i$ is the feature vector of node $i$, $\hat{X}_i$ is the predicted feature vector of node $i$, $E_{ij}$ is the edge feature vector between nodes $i$ and $j$, and $\hat{E}_{ij}$ is the predicted edge feature vector between nodes $i$ and $j$.

The GAE is trained using the pruning environment, where the agent interacts with the environment and collects data to train the GAT encoder.
The actions taken to interact with the environment are randomly sampled from the action space. 
This way, the GAT encoder learns to reconstruct the graph representation of the target neural network while the agent learns to prune the network.
The optimizer used to train the GAE is Adam \cite{Kingma2014-ch}, which is a popular optimizer for training neural networks.
This pre-training step can significantly improve the training stability of the agent, as the GAT encoder has already learned a meaningful representation of the graph representation of the target neural network.
In our experiments, we found that pre-training the GAT encoder using the GAE improved convergence speed of the agent and overall performance.
For this, the encoder was pre-trained for 200 random episodes.

%% file: text/results.tex
In this section, we evaluate our proposed graph-based reinforcement learning approach for neural network pruning.
We first present an analysis of the self-competition reward system to demonstrate its effectiveness in guiding the agent towards optimal pruning policies.
Next, we analyze the learned pruning policies by comparing them to traditional weight magnitude-based pruning methods.
Finally, we compare the performance of our method with other traditional and AutoML-based pruning methods across various datasets and models.

\subsection{Self-Competition Analysis}

In this section, we analyze the effectiveness of the self-competition reward system. 
For this analysis, we examine the trajectory of the relevant metrics during the training of the agent.
We analyze both operational modes of the self-competition reward system: the resource constrained mode and the performance guaranteed mode.

\subsubsection{Resource Constrained Mode}

\begin{figure*}[ht]
    \centering
    \begin{subfigure}[t]{.49\textwidth}
        \centering
        \includegraphics[width=\linewidth]{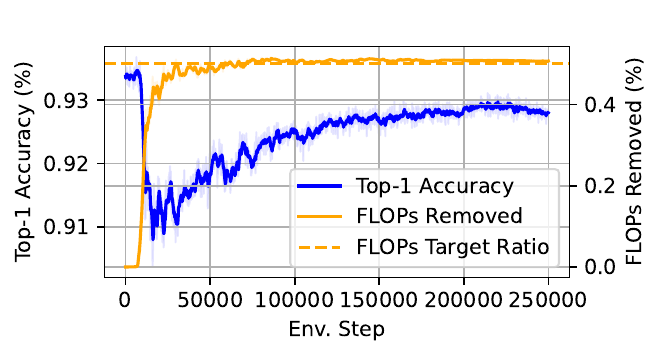}
        \caption{VGG-16 accuracy-sparsity trajectory on CIFAR-10 under FLOPs constraints with a target of 50\% removed FLOPs.}
        \label{fig:self_competition_vgg16_cifar10_0_5}
    \end{subfigure}
    \hfill
    \begin{subfigure}[t]{.49\textwidth}
        \centering
        \includegraphics[width=\linewidth]{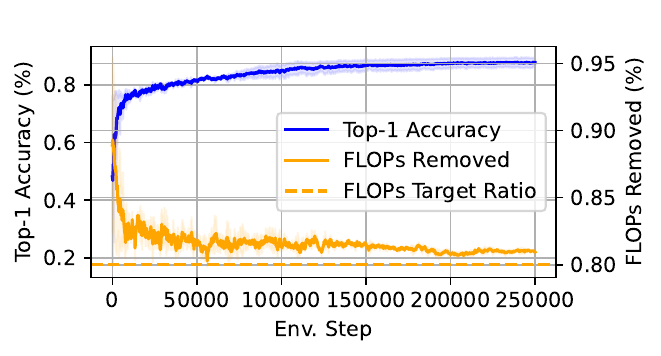}
        \caption{VGG-16 accuracy-sparsity trajectory on CIFAR-10 under FLOPs constraints with a target of 80\% removed FLOPs.}
        \label{fig:self_competition_vgg16_cifar10_0_8}
    \end{subfigure}
    \caption{Trajectories of the agent during training. The graphs show the trajectory of the agent in resource constrained mode. The self-competition reward system first incentivizes the agent to decrease the number of FLOPs until a certain target is reached. Afterward, the agent begins to improve the accuracy again within the constraints of the FLOPs target. Results are shown for the VGG-16 model on CIFAR-10 with different FLOPs targets: 50\% and 80\% removed FLOPs. The environment was configured with $n > 1$.}
    \label{fig:self_competition_trajectory}
\end{figure*}

In Figure \ref{fig:self_competition_trajectory}, we examine the resource constrained mode, where the reward system is designed to maximize the task performance while satisfying the target constraints.
The task performance in our experiments is measured by the top-1 classification accuracy of the target model.
The resource constraints are measured by the ratio of removed FLOPs from the target model.
In the analysis, we show the trajectories of VGG-16 models \cite{Simonyan2014-nx} on CIFAR-10 \cite{Krizhevsky2009-kf}.
The two subfigures \ref{fig:self_competition_vgg16_cifar10_0_5} and \ref{fig:self_competition_vgg16_cifar10_0_8} show the results for different FLOPs targets: 50\% and 80\% removed FLOPs.
It is important to note that the environment was configured with $n > 1$, meaning the agent prunes one group of channels at a time.
In both figures, we can see the evolution of the accuracy and the percentage of FLOPs removed from the target model during training of the agent.
The graph clearly presents the two phases in which the self-competition reward system operates and the agent optimizes its policy.
In the first phase, the agent focuses on achieving the target FLOPs constraint. 
Depending on the policy initialization and the target constraints, the agent can already satisfy the constraints at the beginning of the training.
In Figure \ref{fig:self_competition_vgg16_cifar10_0_5}, the agent starts with a policy that does not satisfy the target constraint and focuses first on reducing the number of FLOPs, regardless of the accuracy drop.
Then in the second phase, the agent begins to improve the accuracy of the model while respecting the target FLOPs constraint.
In Figure \ref{fig:self_competition_vgg16_cifar10_0_8}, the agent starts with a policy that already satisfies the target constraint and immediately begins to improve the accuracy of the model, potentially increasing the FLOPs of the model again within the constraints of the target number of FLOPs.

\subsubsection{Performance Guaranteed Mode}

\begin{figure}[h]
    \centering
    \includegraphics[width=0.49\linewidth]{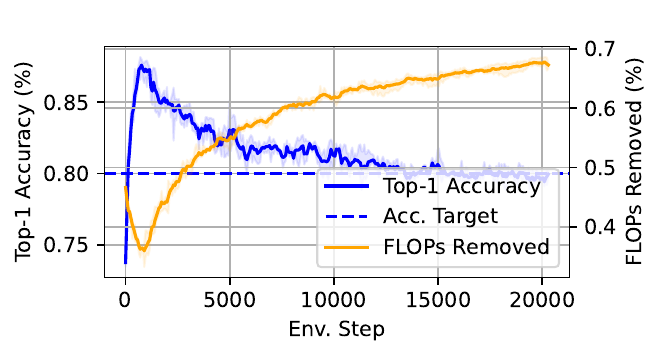}
    \caption{Trajectory of the agent during training. The graphs show the trajectory of the agent in accuracy guarantee mode. In this mode, the agent is incentivized to minimize resource consumption while maintaining a specified accuracy threshold. Results are shown for the ResNet-56 model on CIFAR-10.}
    \label{fig:self_competition_trajectory_acc_guarantee}
\end{figure}

Next, we analyze the performance guaranteed mode shown in Figure \ref{fig:self_competition_trajectory_acc_guarantee}, where the agent aims to minimize resource consumption while maintaining a specified performance threshold.
Figure \ref{fig:self_competition_trajectory_acc_guarantee} shows the trajectory for ResNet-56 on CIFAR-10.
For this experiment the environment was configured with $n = 1$, meaning the agent prunes the entire model at once.
We observe that the agent first focuses on achieving and maintaining the accuracy above the target threshold, this is the first phase of the self-competition reward system.
Once this is achieved, it gradually reduces the FLOPs while ensuring the accuracy doesn't drop below the threshold, in the second phase.
This behavior demonstrates the effectiveness of the reward signal in balancing the trade-off between resource efficiency and maintaining performance guarantees.
The trajectories show that the agent successfully explores pruning configurations while respecting the accuracy constraint, ultimately finding efficient network architectures that meet both the performance requirements and minimize resource usage.

Our results confirm the effectiveness of the self-competition reward system in both modes: prioritizing compression before accuracy in resource constrained mode, and maintaining accuracy while improving efficiency in performance guaranteed mode.

While our method achieves competitive accuracy, there exists a trade-off between compression and performance. 
Our approach effectively minimizes this trade-off through the self-competition reward system, which balances these competing objectives. 
The flexibility of this approach allows it to be adapted to different optimization objectives while maintaining stable and effective performance. 
As shown in our experimental results, higher compression rates typically lead to some accuracy degradation, but our method manages this trade-off better than traditional approaches by learning model-specific pruning strategies.

\subsection{Policy Analysis}

In this section, we analyze the learned policies of the agent.
We do this by comparing them to traditional weight magnitude-based pruning policies.
Specifically, we compare our learned policies to the weight-based pruning policies described in \cite{Balemans2025-lx}.
In that work, the mask is determined by the weight magnitudes of the channels, where the channels with the smallest weight magnitudes are pruned.
The agent does not learn the mask directly, but rather learns a pruning rate for each layer which in combination with the weight magnitudes determines the pruning mask.
We perform a qualitative analysis by visualizing the pruning masks and comparing the masks of the learned policies to the weight-based pruning masks.
This analysis allows us to identify which layers and, more specifically, which channels are pruned by each method.
Additionally, we perform a quantitative analysis by comparing statistics of the pruning results and evaluating the similarity of the masks.
We compare the masks using three metrics: the Jaccard Index, the Cosine Similarity, and the Hamming Distance. 
The Jaccard Index measures the similarity between two sets by dividing the size of their intersection by the size of their union. The Cosine Similarity measures the cosine of the angle between two non-zero vectors, which is a measure of orientation rather than magnitude. The Hamming Distance measures the number of positions at which the corresponding elements of two binary vectors are different. For the Jaccard Index and Cosine Similarity, a higher value indicates greater similarity, while for the Hamming Distance, a lower value indicates greater similarity.
Together, these metrics provide a comprehensive view of the similarity between the learned and weight-based pruning masks.
These statistics are compared on a layer-by-layer basis, where we measure the sparsity (determined by the number of FLOPs before and after pruning) and the number of channels pruned per layer.
We analyze the policies of two different models: VGG-16 and ResNet-56, both trained on CIFAR-10.
Both these models are pruned with the self-competition reward system until convergence.

\input{results/grouped_figures/vgg_masks.tex}

For VGG-16, the model is pruned using the resource constrained mode with a target of 50\% removed FLOPs.
At this target, the traditional weight magnitude pruning method achieves a top-1 accuracy of 91.76\% on CIFAR-10, while our graph-based learned mask method achieves a top-1 accuracy of 92.41\%.
This means that at the same resource budget our learned mask achieves a slightly better accuracy than the traditional weight magnitude pruning method.
The difference for this can be attributed to the flexibility of the learned mask policy, which allows it to adaptively prune channels based on their importance for the task, rather than relying solely on weight magnitudes.
This can be seen in the pruning masks shown in Figure \ref{fig:masks_vgg_cifar_0_5_weight_based} and \ref{fig:masks_vgg_cifar_0_5_learned}.
In these masks we see similarities in the pruning patterns, particularly in the early layers of the network, where both methods tend to prune similar channels.
The learned mask method, however, shows a more adaptive pruning strategy, especially in deeper layers, where it makes different, task-specific choices that are not solely dependent on weight magnitude.
This is also apparent in the similarity metrics shown in Figure \ref{fig:policy_mask_metrics_vgg_cifar_0_5}.
The earlier layers show a higher similarity between the learned mask and the weight-based mask.
This can be explained by the fact that the earlier layers tend to encode more general features, which are often less task-specific.
In the deeper layers we see more differences, the sparsity ratios close to each other, however we see the similarities are lower, indicating that the learned mask is able to adaptively prune channels based on their importance for the task.
The fully learned policies with layer-wise channel count and sparsity ratios are shown in Figure \ref{fig:pruning_policy_combo_vgg_cifar_0_5_weight_based} and Figure \ref{fig:pruning_policy_combo_vgg_cifar_0_5_learned}.
These figures further illustrate the observed trends, where the learned mask method shows a more adaptive pruning strategy, particularly in deeper layers.
In summary, the observed differences in pruning policies indicate that our learned mask method is able to identify and prune channels that are less critical for the task, leading to a better performing model at the same resource budget.
The analysis indicates that while some overlap exists, our learned policy does not simply replicate magnitude pruning.

\input{results/grouped_figures/resnet_masks.tex}

For ResNet-56, we used the performance guaranteed mode with a target accuracy of 80\%.
At this target the traditional weight magnitude pruning method achieves a FLOPs reduction of 52\%, while our learned mask method achieves a FLOPs reduction of 69\%.
This significant difference demonstrates the learned mask method's superior ability to identify less critical channels while maintaining the target accuracy.
The pruning masks for ResNet-56 are shown in Figure \ref{fig:masks_resnet_cifar_0_8_weight_based} and \ref{fig:masks_resnet_cifar_0_8_learned}.
The similarity metrics shown in Figure \ref{fig:policy_mask_metrics_resnet_cifar_0_8} and full network policies are shown in Figure \ref{fig:pruning_policy_combo_resnet_cifar_0_8_weight_based} and \ref{fig:pruning_policy_combo_resnet_cifar_0_8_learned}.

We attribute the larger differences observed for this model to the graph-based method's substantially higher compression rate. 
Figures \ref{fig:pruning_policy_combo_resnet_cifar_0_8_weight_based} and \ref{fig:pruning_policy_combo_resnet_cifar_0_8_learned} show that while the pruning policy per layer follows a similar pattern with three distinct groups or blocks of layers, the learned mask method prunes much more aggressively and less uniformly within these groups.
This adaptive behavior, also reflected in the lower similarity scores compared to the VGG-16 model, indicates that our graph-based method identifies channels based on task-specific importance rather than weight magnitudes alone.

Visual analysis of the pruning masks in Figure \ref{fig:masks_resnet_cifar_0_8} confirms these findings.
The learned mask method exhibits more irregular patterns and prunes more channels overall compared to the weight-based approach.
While similarities exist in layer-wise pruning patterns, the learned mask demonstrates higher irregularity, indicating a more flexible pruning strategy that removes functionally redundant connections beyond simple magnitude ordering.
Overall, these results demonstrate that our graph-based approach significantly outperforms traditional magnitude-based pruning for ResNet-56, achieving 17\% better compression efficiency while maintaining the same task accuracy.

To conclude, our analysis of the learned policies reveals several key insights about the effectiveness of our graph-based pruning approach compared to traditional weight magnitude-based methods.
Our learned policies consistently outperform magnitude-based pruning across different architectures and optimization objectives, demonstrating superior compression efficiency while maintaining task performance. 
The similarity analysis reveals that our learned policies do not simply replicate magnitude-based pruning strategies. While some overlap exists, particularly in early layers that encode general features, our approach identifies distinct sets of channels for pruning in deeper layers responsible for task-specific feature extraction. This adaptive behavior indicates that channel importance for a given task extends beyond simple weight magnitude considerations.
The visual analysis of pruning masks further supports these findings, showing that our learned policies exhibit more irregular and flexible pruning patterns compared to the uniform patterns characteristic of magnitude-based methods.
This irregularity demonstrates that our approach successfully learns to identify functionally redundant connections that may not correspond to the smallest absolute weight values.
Collectively, these findings demonstrate that our graph-based approach learns task-specific pruning strategies that achieve superior compression efficiency while better preserving model performance. The ability to adaptively identify channel importance based on task requirements, rather than relying solely on weight statistics, represents a significant advancement in neural network pruning methodology.

\subsection{The impact of $n$} \label{subsec:n_impact} 

The parameter $n$ controls the size of the channel groups in the action space and, by extension, the number of environment interactions available to the agent within the self-competition reward framework. When $n=1$, the agent performs a single, global pruning step. For $n>1$, pruning proceeds sequentially over multiple steps, each targeting one group of channels, and the environment state is updated after each action.

In our experiments, we observe consistently faster convergence (i.e., fewer environment interactions to reach comparable performance) for $n=1$ across models and datasets. This effect is expected: the agent optimizes a single pruning mask without learning a temporal policy. As $n$ increases, the agent must solve a longer-horizon credit-assignment problem, which slows convergence, although final performance remains similar in our experiments.

From a representational standpoint, however, the choice of $n$ has important implications for the learned encoder. With $n=1$, the Graph Attention Network (GAT) encoder is trained to produce embeddings from the initial state alone, which accelerates learning but may limit the emergence of state representations that generalize across architectures or pruning objectives. In contrast, for $n>1$ the agent processes sequences of actions together with updated states; this sequential conditioning forces the GAT encoder to capture the evolving network structure and interdependencies between pruned components, potentially yielding more informative and transferable embeddings. The cost of this representational advantage is slower convergence. Improving this training dynamic (i.e., retaining the generalization benefits of trajectory-conditioned embeddings while mitigating the sample-efficiency penalty) and evaluating the GAT's generalizing capability are key directions for future work.

In most of our experiments, we used $n = 4$ to balance convergence speed and the ability to learn a sequence of actions.

\subsection{Comparison with evolutionary search strategies} \label{subsec:graph:n_1_vs_evolutionary_search}

When $n=1$, the agent prunes the entire model in a single step, effectively reducing the problem to a combinatorial optimization task.
In this case, the agent's task is to learn a direct mapping from the initial state of the network to a binary pruning mask that satisfies the resource constraints while maximizing performance. This setup is similar to an evolutionary search over the space of possible pruning masks.
Therefore, we compare the performance of our $n=1$ agent to an evolutionary search baseline that optimizes binary pruning masks using the evolutionary strategies (ES) algorithm \cite{Tripathy1982-kn}.
More specifically, we use the Covariance Matrix Adaptation Evolution Strategy (CMA-ES) \cite{Hansen2001-nl, Hansen2016-aw}, which is a popular variant of ES that adapts the covariance matrix of the search distribution to improve convergence.
We use the same pruning environment, however, as the self-competition reward system is not applicable in this case, we use a reward signal that is based on the task performance and the resource constraints.
The reward signal is defined as follows:
\begin{equation}
    R = 
    \begin{cases}
        \text{Acc}_{\text{top1}} & \text{if } S > S_{\text{target}} \\
        \text{Acc}_{\text{top1}} - |S - S_{\text{target}}| & \text{if } S \leq S_{\text{target}}
    \end{cases}
\end{equation}
where $S = \frac{F}{F_{\text{original}}}$ represents the relative FLOPs ratio.\\
This function penalizes the agent for not satisfying the resource constraints, while still incentivizing the agent to maximize the task performance.
However, this reward signal does not provide a curriculum for the agent to follow, as the self-competition reward system does.
We compare the performance of the $n=1$ agent to the evolutionary search baseline on the same models and datasets as in Section \ref{subsec:comparison_with_other_pruning_methods}.
Both algorithms are run until convergence.
The results are presented in Table \ref{table:graph:combinatorial_accuracy}, where we show the top-1 accuracy without fine-tuning, the difference in top-1 accuracy compared to the baseline model, and the compression setting used to produce the compressed model.

\input{results/tables/combinatorial_table.tex}

From the results, we can see that the CMA-ES algorithm achieves competitive accuracy compared to our $n=1$ agent for the smaller models on the CIFAR-10 dataset. For the larger models and datasets, however, the performance of the CMA-ES algorithm degrades significantly.
We attribute this to the fact that the CMA-ES algorithm is not able to effectively explore the large search space of possible pruning masks for larger models.
We did not include a comparison on the ImageNet dataset as the CMA-ES algorithm was not able to converge within a reasonable time frame, or did not adhere to the constraints. This can be attributed to the used reward function. However, tuning the reward function to improve the performance of the CMA-ES algorithm is out of the scope of this work.
Overall, the results demonstrate that our $n=1$ agent is able to effectively learn a pruning policy that is competitive and outperforms the evolutionary search baseline for larger models and datasets.

\subsection{Comparison with Other Pruning Methods} \label{subsec:comparison_with_other_pruning_methods}

In this section, we compare the results of our method with other traditional and AutoML-based pruning methods.
We use standard models trained on common computer vision classification datasets including CIFAR-10, CIFAR-100 \cite{Krizhevsky2009-kf}, and ImageNet \cite{Deng2009-qa}.

The methods include traditional pruning methods such as Structured Probabilistic Pruning (SPP) by Wang et al. \cite{Wang2017-od}, Filter Pruning (FP) by Li et al. \cite{Li2016-ss}, Soft Filter Pruning (SFP)
by He et al. \cite{He2018-mk}, FPGM by He et al. \cite{He2018-fl}, DSA by Ning et al. \cite{Ning2020-yu}, and Provable Filter Pruning (PFP) by Liebenwein et al. \cite{Liebenwein2019-xl}.

The AutoML methods include NetAdapt by Yang et al. \cite{Yang2018-dz}, AutoPruner by Luo et al. \cite{Luo2018-qy}, EagleEye by Li et al. \cite{Li2020-yq}, Meta-Pruning by Liu et al. \cite{Liu2019-aa}, AutoSlim by Yu et al. \cite{Yu2019-ar}, Runtime Pruning (RNP) by Lin et al. \cite{Lin2017-kw}, Automatic Group-based Structured Pruning (AGSPRL) by Wei et al. \cite{Wei2022-fb}, Constrained Reinforcement Learning (CRL) by Malik et al. \cite{Malik2021-oe}, and AMC by He et al. \cite{He2018-uv}. We refer to our method as Graph-based Self-Competition Compression (G-SCC) in the tables.

\input{results/tables/cifar10_table}

\input{results/tables/cifar100_table}

\input{results/tables/imagenet_table}

The results are presented in Table \ref{table:graph:cifar10_accuracy} for the CIFAR-10 dataset, Table \ref{table:graph:cifar100_accuracy} for the CIFAR-100 dataset, and Table \ref{table:graph:imagenet_accuracy} for the ImageNet dataset.
The tables display the top-1 accuracy, the difference in top-1 accuracy compared to the baseline model, and the compression setting used to produce the compressed model.
The compression setting is defined by the target compression ratio, which is measured in terms of FLOPs or parameter count.
It is crucial to highlight that post-compression, the models are fine-tuned to further minimize the loss in accuracy.
The fine-tuning process allows the models to recover some of the accuracy lost during pruning, which is a common practice in neural network compression.
The models were fine-tuned for 128 epochs on CIFAR-10/100 and 64 epochs on ImageNet using the Adam optimizer with a smaller learning rate of $10^{-4}$, which is decayed based on a reduce-on-plateau scheduler that monitors the validation accuracy.
Minor discrepancies in the reported accuracy values can be attributed to the fine-tuning process and the specific implementation details inherent to each method. 
Not all methods disclose their fine-tuning process, potentially leading to variations in the reported accuracy values. 
Moreover, we observed variations in the reported baseline accuracy models across different studies.
Therefore, we include both the raw pruned top-1 accuracy and the fine-tuned top-1 accuracy for the models that we evaluated ourselves.

The results demonstrate that our method achieves competitive accuracy across all datasets while maintaining a high compression rate.
For most models we see our method maintains higher accuracy before fine-tuning compared to the other methods.
This means that our method is able to identify and prune less critical channels more effectively, leading to a more efficient model.
After fine-tuning, the differences in accuracy are less pronounced, which is expected as the fine-tuning process allows the models to recover some of the accuracy lost during pruning.
Overall, our method demonstrates a strong performance in terms of both compression rate and accuracy across multiple datasets and models.
In certain instances, there is even an observed increase in accuracy compared to the original baseline model. 
We attribute this improvement to the regularization effect of the pruning process, which can aid in mitigating overfitting and enhancing generalization \cite{Zhu2017-gs} \cite{Liu2017-zi}.
Overall, the results demonstrate that our approach effectively achieves the desired compression rate while maintaining competitive accuracy performance.

%% file: results/grouped_figures/vgg_masks.tex
\begin{figure*}[h!]
    \centering
    
    \begin{subfigure}[t]{.49\textwidth}
        \centering
        \includegraphics[width=\linewidth]{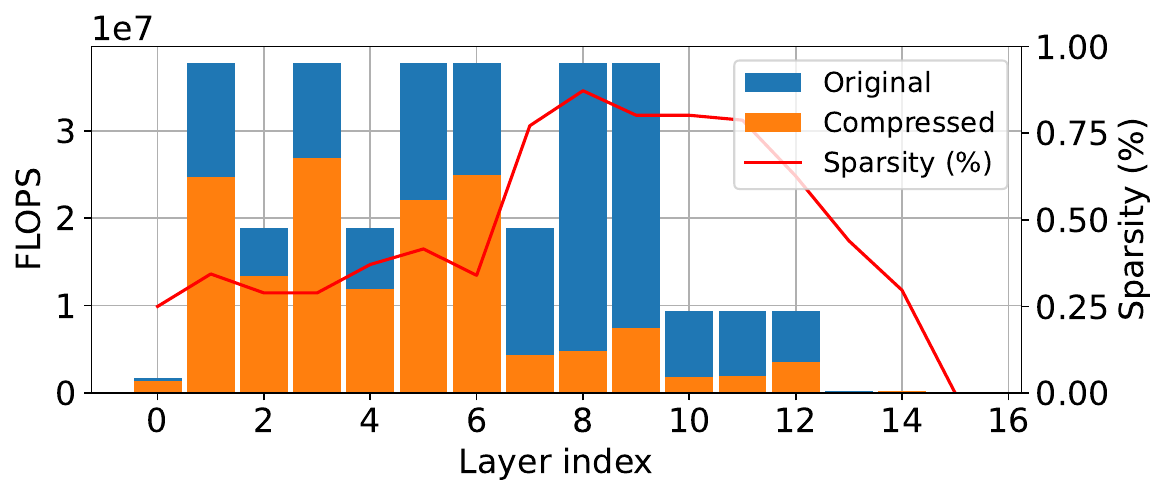}
        \caption{Pruning policy of VGG-16 on CIFAR-10 with 50\% removed FLOPs using weight-based pruning mask.}
        \label{fig:pruning_policy_combo_vgg_cifar_0_5_weight_based}
    \end{subfigure}
    \hfill
    \begin{subfigure}[t]{.49\textwidth}
        \centering
        \includegraphics[width=\linewidth]{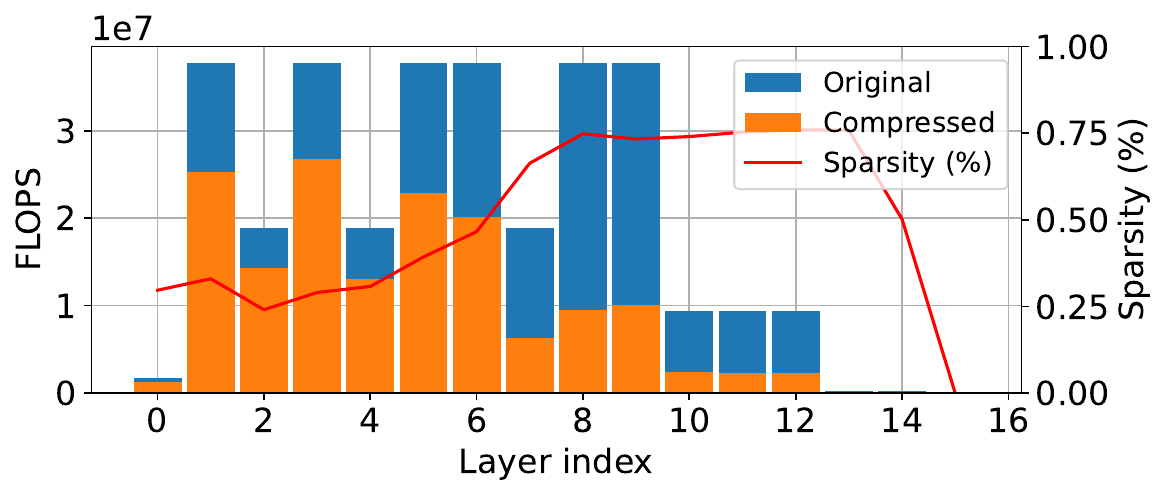}
        \caption{Pruning policy of VGG-16 on CIFAR-10 with 50\% removed FLOPs using learned pruning mask (ours).}
        \label{fig:pruning_policy_combo_vgg_cifar_0_5_learned}
    \end{subfigure}
    
    \vspace{0.3cm}

    \begin{subfigure}[t]{\linewidth}
        \centering
        \includegraphics[width=\linewidth]{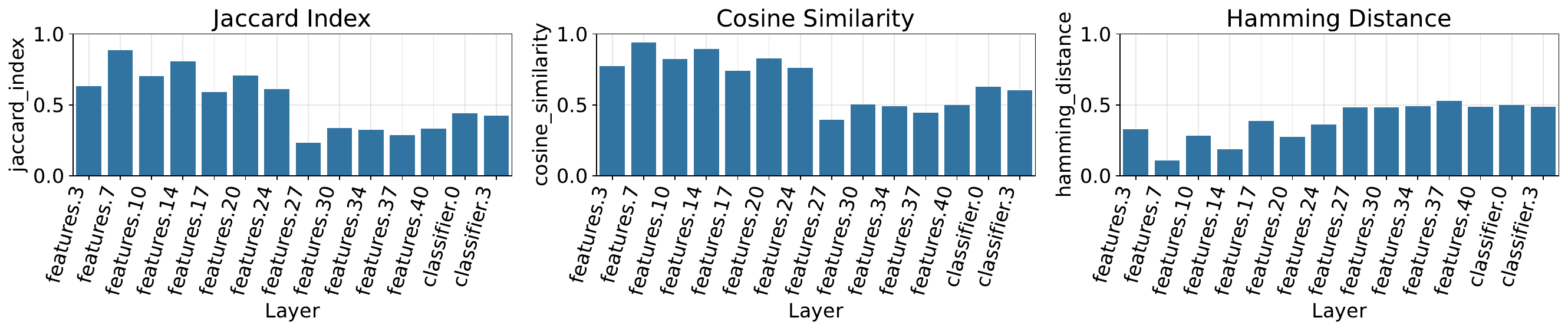}
        \caption{Similarity metrics of the pruning masks for VGG-16 on CIFAR-10 with 50\% removed FLOPs. The similarity is measured by the Jaccard index, the cosine similarity, and the Hamming distance. The metrics are calculated on a layer-by-layer basis, comparing the learned pruning mask to the weight-based pruning mask.}
        \label{fig:policy_mask_metrics_vgg_cifar_0_5}
    \end{subfigure}
    
    \vspace{0.3cm}

    \begin{subfigure}[t]{.95\textwidth}
        \centering
        \includegraphics[width=\linewidth]{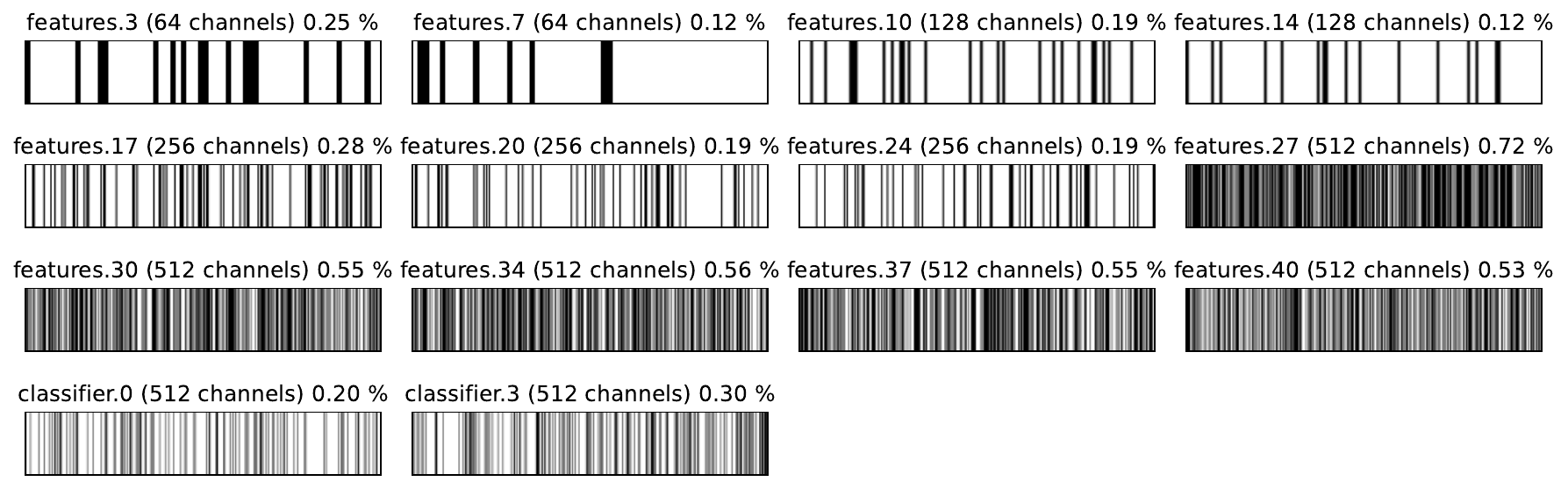}
        \caption{Pruning masks of VGG-16 on CIFAR-10 with 50\% removed FLOPs using weight-based pruning mask.}
        \label{fig:masks_vgg_cifar_0_5_weight_based}
    \end{subfigure}
    \hfill
    \begin{subfigure}[t]{.95\textwidth}
        \centering
        \includegraphics[width=\linewidth]{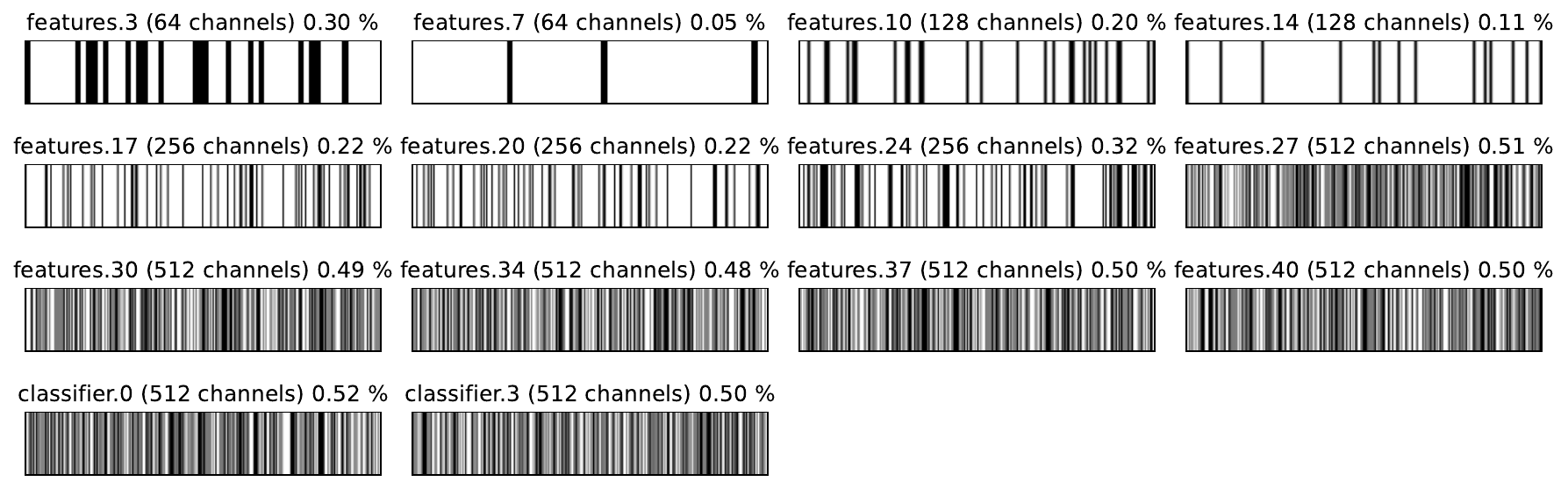}
        \caption{Pruning masks of VGG-16 on CIFAR-10 with 50\% removed FLOPs using learned pruning mask (ours).}
        \label{fig:masks_vgg_cifar_0_5_learned}
    \end{subfigure}
    
    \caption{Analysis of pruning policies for VGG-16 on CIFAR-10 with 50\% removed FLOPs. (a-b) Comparison of pruning policies showing layer-wise channel count and sparsity ratios. (c) Similarity metrics comparing learned and weight-based masks. (d-e) Visual representation of the actual pruning masks applied to each layer.}
    \label{fig:vgg_cifar_policy_analysis}
\end{figure*}

%% file: results/grouped_figures/resnet_masks.tex
\begin{figure*}[h]
    \centering
    \begin{subfigure}[t]{.49\textwidth}
        \centering
        \includegraphics[width=\linewidth]{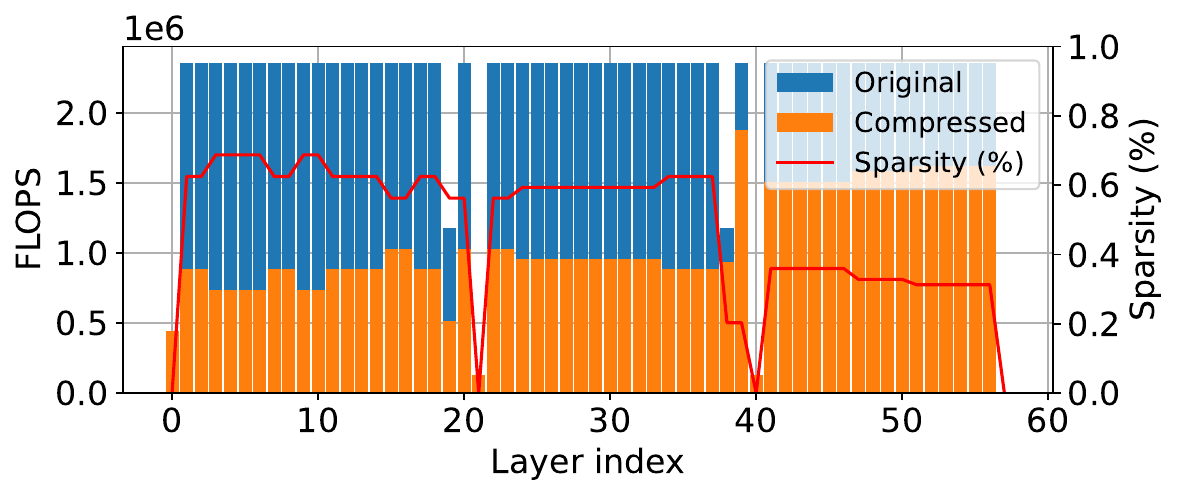}
        \caption{Pruning policy of ResNet-56 on CIFAR-10 with 80\% accuracy guaranteed using weight-based pruning mask.}
        \label{fig:pruning_policy_combo_resnet_cifar_0_8_weight_based}
    \end{subfigure}
    \hfill
    \begin{subfigure}[t]{.49\textwidth}
        \centering
        \includegraphics[width=\linewidth]{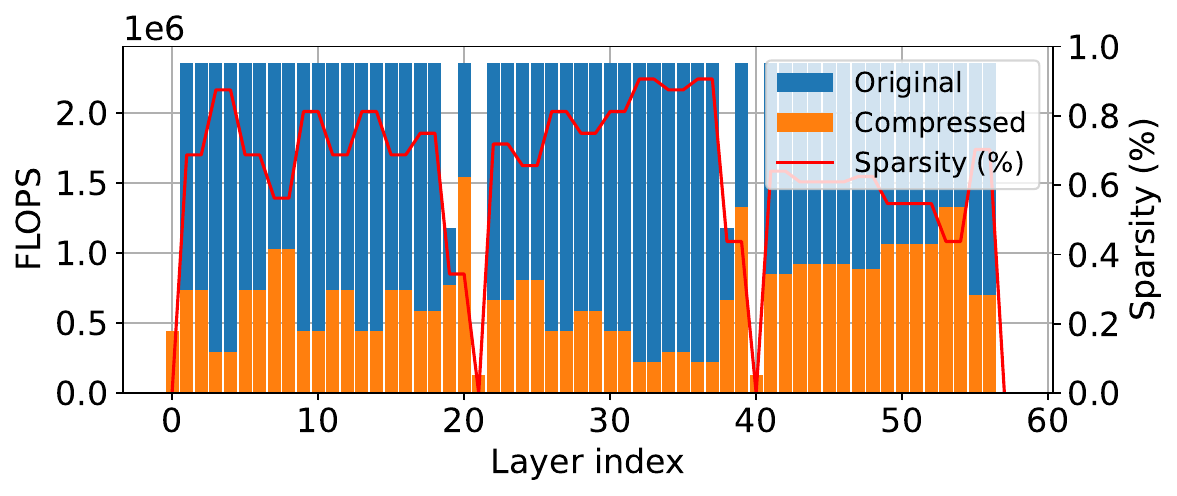}
        \caption{Pruning policy of ResNet-56 on CIFAR-10 with 80\% accuracy guaranteed using learned pruning mask (ours).}
        \label{fig:pruning_policy_combo_resnet_cifar_0_8_learned}
    \end{subfigure}
    
    \vspace{0.3cm}

    \begin{subfigure}[t]{\linewidth}
        \centering
        \includegraphics[width=\linewidth]{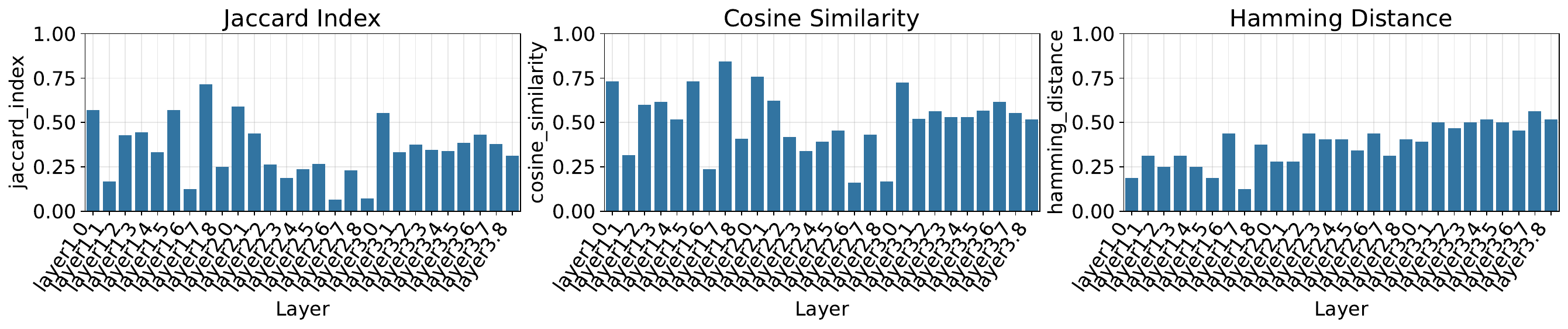}
        \caption{Similarity metrics of the pruning masks for ResNet-56 on CIFAR-10 with 80\% accuracy guaranteed. The similarity is measured by the Jaccard index, the cosine similarity, and the Hamming distance. The metrics are calculated on a layer-by-layer basis, comparing the learned pruning mask to the weight-based pruning mask.}
        \label{fig:policy_mask_metrics_resnet_cifar_0_8}
    \end{subfigure}
    
    \caption{Analysis of pruning policies for ResNet-56 on CIFAR-10 with 80\% accuracy guaranteed. (a-b) Comparison of pruning policies showing layer-wise channel count and sparsity ratios. (c) Similarity metrics comparing learned and weight-based masks.}
    \label{fig:resnet_cifar_policy_analysis}
\end{figure*}

\begin{figure*}[p]
    \centering
    
    \begin{subfigure}[t]{.95\textwidth}
        \centering
        \includegraphics[width=\linewidth]{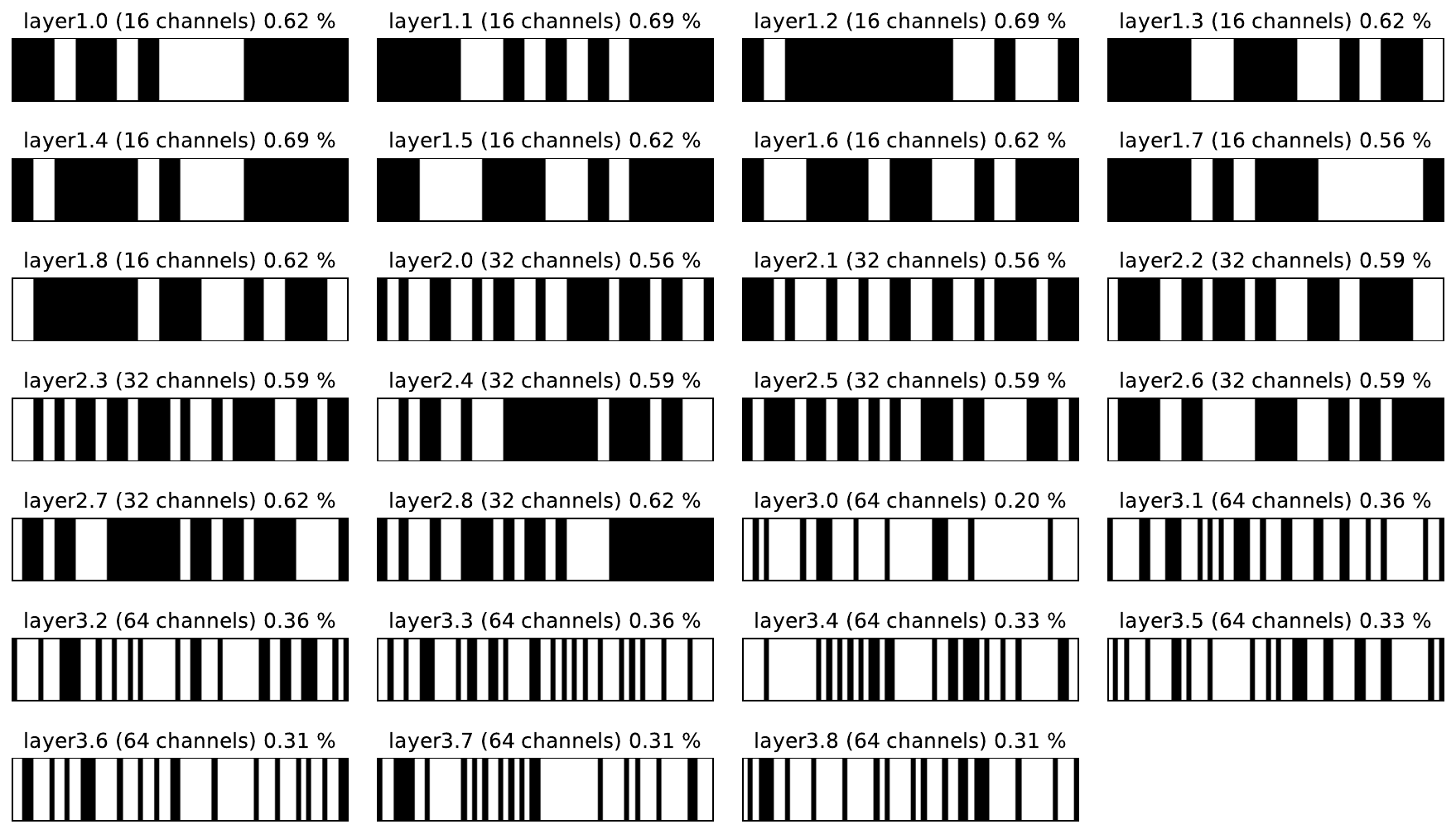}
        \caption{Pruning masks of ResNet-56 on CIFAR-10 with 80\% accuracy guaranteed using weight-based pruning mask.}
        \label{fig:masks_resnet_cifar_0_8_weight_based}
    \end{subfigure}
    
    \vspace{0.3cm}
    
    \begin{subfigure}[t]{.95\textwidth}
        \centering
        \includegraphics[width=\linewidth]{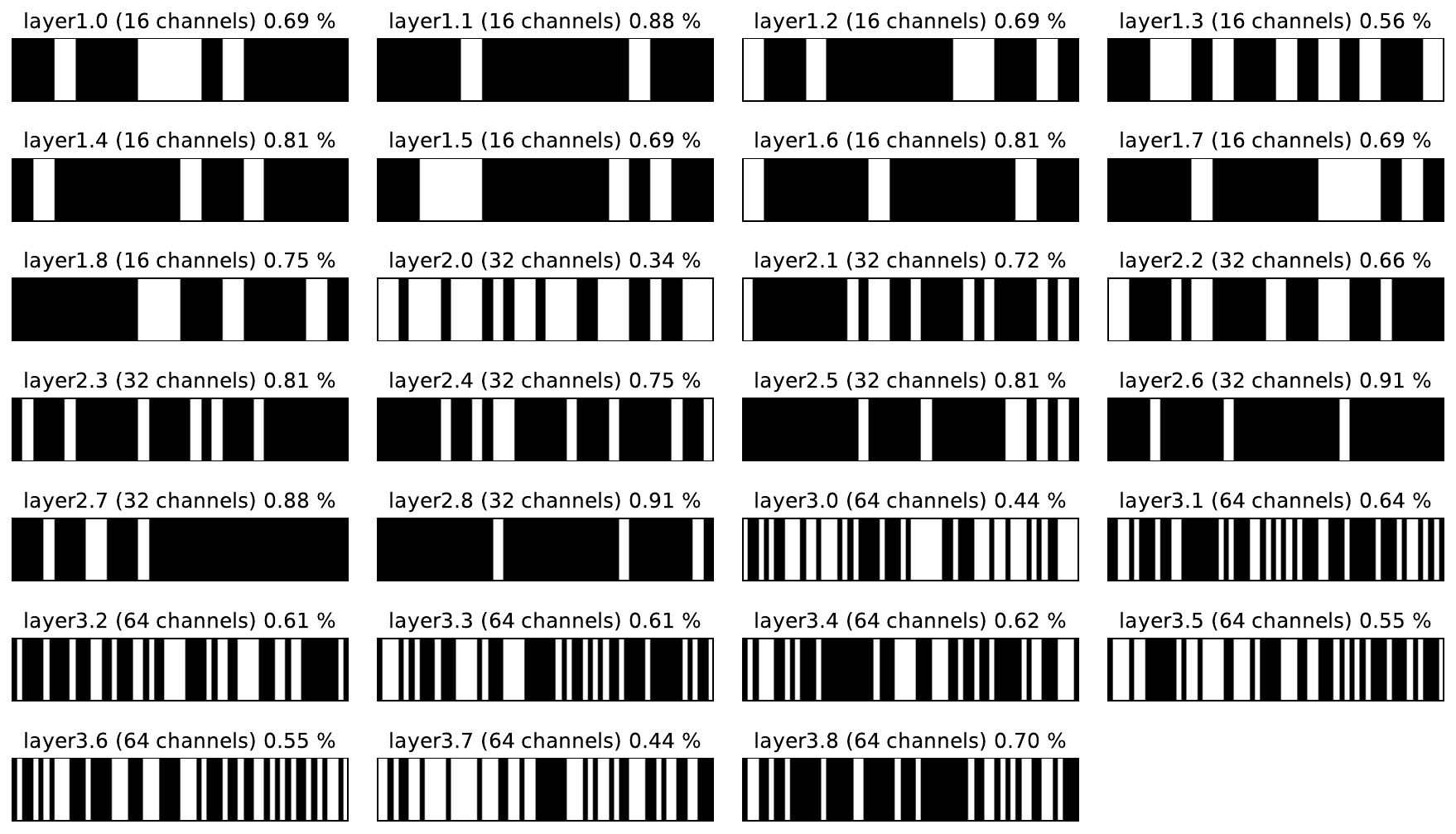}
        \caption{Pruning masks of ResNet-56 on CIFAR-10 with 80\% accuracy guaranteed using learned pruning mask (ours).}
        \label{fig:masks_resnet_cifar_0_8_learned}
    \end{subfigure}
    \caption{Pruning masks of ResNet-56 on CIFAR-10 with 80\% accuracy guaranteed. (a-b) Visual representation of the actual pruning masks applied to each layer.}
    \label{fig:masks_resnet_cifar_0_8}
\end{figure*}

%% file: results/tables/combinatorial_table.tex
\begin{table}[t]
\centering
\resizebox{\columnwidth}{!}{
\begin{tabular}{l l l l r r}
\hline
\textbf{Model} & \textbf{Dataset} & \textbf{Method} & \textbf{Ratio} & \textbf{Top-1 (\%)} & \textbf{$\Delta$Top-1 (\%)} \\
\hline
\multirow{4}{*}{VGG-16} & \multirow{2}{*}{Cifar-10} & G-SCC $(n=1)$ (ours) & 50\% FLOPs & 92.69 & -0.93 \\
 & & CMA-ES & 50\% FLOPs & 93.55 & -0.07 \\
\cline{2-6}
 & \multirow{2}{*}{Cifar-100} & G-SCC $(n=1)$ (ours) & 40\% FLOPs & 71.51 & -2.49 \\
 & & CMA-ES & 50\% FLOPs & 65.62 & -8.38 \\
\hline
\multirow{2}{*}{ResNet-56} & \multirow{2}{*}{Cifar-10} & G-SCC $(n=1)$ (ours) & 50\% FLOPs & 90.21 & -4.16 \\
 & & CMA-ES & 50\% FLOPs & 90.57 & -3.80 \\
\hline
\multirow{4}{*}{Mobilenet-V2} & \multirow{2}{*}{Cifar-10} & G-SCC $(n=1)$ (ours) & 50\% FLOPs & 92.88 & -0.73 \\
 & & CMA-ES & 50\% FLOPs & 89.48 & -4.13 \\
\cline{2-6}
 & \multirow{2}{*}{Cifar-100} & G-SCC $(n=1)$ (ours) & 50\% FLOPs & 74.89 & +0.69 \\
 & & CMA-ES & 50\% FLOPs & 64.63 & -9.57 \\
\hline
\end{tabular}
}
\caption[Combinatorial results on CIFAR-10 and CIFAR-100.]{Results of comparing our methods versus Evolutionary Strategies on different models trained on CIFAR-10 and CIFAR-100. The table shows the dataset, top-1 accuracy after pruning without fine-tuning, the difference to the baseline model, and the compression setting used to produce the compressed the model.}
\label{table:graph:combinatorial_accuracy}
\end{table}

%% file: results/tables/cifar10_table.tex
\begin{table}[h]
\centering
\begin{tabular}{l l l r r}
\hline
\textbf{Model} & \textbf{Method} & \textbf{Ratio} & \textbf{Top-1 (\%)} & \textbf{$\Delta$Top-1 (\%)} \\
\hline
\multirow{6}{*}{VGG-16} & AGSPRL \cite{Wei2022-fb} & 50\% FLOPs & 93.04 & -0.37 \\
 & CRL \cite{Malik2021-oe} & 20\% FLOPs & 90.96 & +0.27 \\
 & SCC \cite{Balemans2025-lx} & 25\% FLOPs & 91.34 (86.83) & -2.28 \\
 & SCC \cite{Balemans2025-lx} & 50\% FLOPs & 93.14 (91.76) & -0.48 \\
 & G-SCC $(n=1)$ (ours) & 50\% FLOPs & 93.67 (92.69) & +0.05 \\
 & G-SCC $(n>1)$ (ours) & 50\% FLOPs & 93.58 (93.55) & -0.04 \\
\hline
\multirow{7}{*}{ResNet-56} & Uniform & 50\% FLOPs & 87.50 & -5.89 \\
 & AMC \cite{He2018-uv} & 50\% FLOPs & 90.20 & -3.19 \\
 & FPGM \cite{He2018-fl} & 48\% FLOPs & 93.26 & -0.33 \\
 & EagleEye \cite{Li2020-yq} & 50\% FLOPs & 94.66 & +1.40 \\
 & SCC \cite{Balemans2025-lx} & 50\% FLOPs & 92.96 (81.72) & -1.41 \\
 & G-SCC $(n=1)$ (ours) & 50\% FLOPs & 93.39 (90.21) & -0.98 \\
 & G-SCC $(n>1)$ (ours) & 50\% FLOPs & 93.25 (86.89) & -1.12 \\
\hline
\multirow{3}{*}{Mobilenet-V2} & SCC \cite{Balemans2025-lx} & 46\% FLOPs & 93.96 (93.33) & +0.35 \\
 & G-SCC $(n=1)$ (ours) & 50\% FLOPs & 93.48 (92.88) & -0.13 \\
 & G-SCC $(n>1)$ (ours) & 50\% FLOPs & 93.65 (92.28) & +0.04 \\
\hline
\end{tabular}
\caption{Results of different methods and models on CIFAR-10. The table shows the top-1 accuracy, the difference to the baseline model, and the compression setting used to produce the compressed the model. For the models that we fine-tuned, the accuracy of the raw compressed model without fine-tuning is given in parentheses.}
\label{table:graph:cifar10_accuracy}
\end{table}

%% file: results/tables/cifar100_table.tex
\begin{table}[h]
\centering
\begin{tabular}{l l l r r}
\hline
\textbf{Model} & \textbf{Method} & \textbf{Ratio} & \textbf{Top-1 (\%)} & \textbf{$\Delta$Top-1 (\%)} \\
\hline
\multirow{4}{*}{VGG-16} & AGSPRL \cite{Wei2022-fb} & 70\% FLOPs & 72.20 & -0.04 \\
 & SCC \cite{Balemans2025-lx} & 50\% FLOPs & 72.59 (67.90) & -1.41 \\
 & G-SCC $(n=1)$ (ours) & 40\% FLOPs & 72.71 (71.51) & -1.29 \\
 & G-SCC $(n>1)$ (ours) & 40\% FLOPs & 71.53 (66.40) & -2.47 \\
\hline
\multirow{5}{*}{ResNet-56} & AGSPRL \cite{Wei2022-fb} & 60\% FLOPs & 72.49 & -0.05 \\
 & SCC \cite{Balemans2025-lx} & 50\% FLOPs & 70.96 (36.64) & -1.67 \\
 & SCC \cite{Balemans2025-lx} & 60\% FLOPs & 72.09 (55.07) & -0.54 \\
 & G-SCC $(n=1)$(ours) & 50\% FLOPs & 70.97 (49.04) & -1.66 \\
 & G-SCC $(n=1)$(ours) & 60\% FLOPs & 71.58 (63.23) & -1.05 \\
\hline
\multirow{2}{*}{Mobilenet-V2} & SCC \cite{Balemans2025-lx} & 50\% FLOPs & 74.07 (72.18) & -0.13 \\
 & G-SCC $(n=1)$(ours) & 50\% FLOPs & 74.89 (74.89) & +0.69 \\
\hline
\end{tabular}
\caption{Results of different methods and models on CIFAR-100. The table shows the top-1 accuracy, the difference to the baseline model, and the compression setting used to produce the compressed the model. For the models that we fine-tuned, the accuracy of the raw compressed model without fine-tuning is given in parentheses.}
\label{table:graph:cifar100_accuracy}
\end{table}

%% file: results/tables/imagenet_table.tex
\begin{table}[ht]
\centering
\begin{tabular}{l l l r r}
\hline
\textbf{Model} & \textbf{Method} & \textbf{Ratio} & \textbf{Top-1 (\%)} & \textbf{$\Delta$Top-1 (\%)} \\
\hline
\multirow{9}{*}{\rotatebox{0}{\parbox{2cm}{\raggedright VGG-16}}} & FP \cite{Li2016-ss} & 20\% FLOPs & 55.90 & -14.60 \\
& RNP \cite{Lin2017-kw} & 20\% FLOPs & 66.92 & -3.58 \\
& SPP \cite{Wang2017-od} & 20\% FLOPs & 68.20 & -2.30 \\
& AMC \cite{He2018-uv} & 20\% FLOPs & 69.10 & -1.40 \\
& AutoPruner \cite{Luo2018-qy} & 26\% FLOPs & 69.20 & -2.39 \\
& SCC \cite{Balemans2025-lx} & 54\% FLOPs & 72.36 (65.34) & -1.01 \\
& SCC \cite{Balemans2025-lx} & 20\% FLOPs & 69.10 (54.71) & -4.27 \\
& G-SCC $(n=1)$ (ours) & 50\% FLOPs & 73.25 (65.89) & -0.12 \\
& G-SCC $(n=1)$ (ours) & 20\% FLOPs & 70.15 (59.12) & -3.22 \\
\hline
\multirow{6}{*}{\rotatebox{0}{\parbox{2cm}{\raggedright ResNet-50}}} & AutoPruner \cite{Luo2018-qy} & 56\% FLOPs & 73.84 & -2.31 \\
& Meta-Pruning \cite{Liu2019-aa} & 50\% FLOPs & 73.40 & -3.20 \\
& AutoSlim \cite{Yu2019-ar} & 50\% FLOPs & 74.00 & N/A \\
& EagleEye \cite{Li2020-yq} & 50\% FLOPs & 74.20 & N/A \\
& SCC \cite{Balemans2025-lx} & 50\% FLOPs & 73.57 (61.94) & -2.57 \\
& G-SCC $(n=1)$ (ours) & 50\% FLOPs & 74.20 (63.91) & -1.94 \\
\hline
\multirow{6}{*}{\rotatebox{0}{\parbox{2cm}{\raggedright Mobilenet-V2}}} & Meta-Pruning \cite{Liu2019-aa} & 73\% FLOPs & 71.20 & -3.50 \\
& NetAdapt \cite{Yang2018-dz} & 75\% FLOPs & 70.00 & N/A \\
& SCC \cite{Balemans2025-lx} & 70\% FLOPs & 69.42 (59.63) & -2.45 \\
& SCC \cite{Balemans2025-lx} & 50\% FLOPs & 66.33 (28.45) & -5.54 \\
& G-SCC $(n=1)$ (ours) & 70\% FLOPs & 70.35 (63.45) & -1.52 \\
& G-SCC $(n=1)$ (ours) & 50\% FLOPs & 67.66 (31.38) & -4.21 \\
\hline
\end{tabular}
\caption{Results of different methods and models on ImageNet. The table shows the top-1 accuracy, the difference to the baseline model, and the compression setting used to produce the compressed the model. For the models that we fine-tuned, the accuracy of the raw compressed model without fine-tuning is given in parentheses.}
\label{table:graph:imagenet_accuracy}
\end{table}

%% file: text/conclusion-and-future-work.tex
This paper addresses fundamental limitations of traditional neural network pruning methods, specifically their reliance on handcrafted heuristics and local optimization perspectives that limit their effectiveness across diverse architectures and deployment scenarios.
Our work transforms the pruning process by introducing a comprehensive graph representation of neural networks that captures the topological relationships between layers and channels, replacing the limited layer-wise observation space of existing methods with a global view of the network structure.

The core technical innovations of our approach include the development of a Graph Attention Network (GAT) encoder that processes the network's graph representation to generate informative embeddings, eliminating the reliance on handcrafted features.
This GAT encoder dynamically attends to relevant network components, learning which parts of the network topology are most important for making pruning decisions.
Furthermore, our transition from continuous pruning ratios to a fine-grained binary action space enables the reinforcement learning agent to learn data-driven channel importance criteria, moving away from predefined heuristic scoring functions such as $l_1$-norm weights.

Central to our methodology is the self-competition reward system.
Rather than enforcing constraints implicitly in the environment or through complex Lagrangian optimization, our approach models constraints within the reward signal itself, enabling the reinforcement learning agent to learn adaptive policies that balance compression and performance objectives.
The self-competition mechanism allows the agent to continuously improve upon its historical performance while prioritizing the satisfaction of (resource) constraints. Using self-competition, compression can be performed in two operational modes: resource-constrained optimization and performance-guaranteed compression.

Our experimental evaluation demonstrates the effectiveness of the proposed framework across multiple architectures and datasets.
The analysis of learned policies reveals that our approach consistently outperforms traditional weight magnitude-based pruning methods, achieving superior compression efficiency while maintaining competitive accuracy.
The comparison with existing methods shows that our framework achieves state-of-the-art results across the CIFAR-10, CIFAR-100, and ImageNet datasets, often matching or exceeding the performance of both traditional and AutoML-based pruning approaches.
Notably, our method demonstrates the ability to learn task-specific pruning strategies that identify functionally redundant connections beyond what is captured by weight magnitude alone.

While the approaches presented in this chapter represent significant advances in automated neural network compression, several limitations and areas for future research remain.
A fundamental limitation of the current framework is that the learned policies are specific to the model architecture, dataset, and target compression rate used during training.
Future work should explore methodologies for generalizing learned policies across different models and datasets while adapting to varying compression requirements, potentially through meta-learning or transfer learning techniques for pruning policies.
The scalability of our approach presents another area for improvement, particularly for exceptionally large models.
The on-policy nature of the PPO agent requires numerous environment interactions, which can be computationally expensive.
Investigating more sample-efficient training methods, such as off-policy algorithms or model-based reinforcement learning, could substantially reduce the computational overhead.
Furthermore, exploring alternative graph neural network architectures beyond GATs could yield more efficient or expressive representations.
Finally, extending this graph-based methodology to other architectures, such as Large Language Models (LLMs) \cite{Naveed2023-jf} and Vision Transformers (ViTs) \cite{Dosovitskiy2020-ss}, would demonstrate the broader applicability of our framework.

The AutoML approaches presented in this work establish a foundation for generalizable compression frameworks that operate effectively across diverse neural network architectures and application domains without relying on handcrafted heuristics.
The ability to learn adaptive, task-specific pruning strategies represents a significant step toward fully automated neural network optimization, addressing the varied requirements of modern edge AI applications.